# Scene-LSTM: A Model for Human Trajectory Prediction

Huynh Manh, Gita Alaghband
University of Colorado Denver
{huynh.manh, gita.alaghband}@ucdenver.edu

**Abstract**

*We develop a human movement trajectory prediction system that incorporates the scene information (Scene-LSTM) as well as human movement trajectories (Pedestrian movement LSTM) in the prediction process within static crowded scenes. We superimpose a two-level grid structure (scene is divided into grid cells each modeled by a scene-LSTM, which are further divided into smaller sub-grids for finer spatial granularity) and explore common human trajectories occurring in the grid cell (e.g., making a right or left turn onto sidewalks coming out of an alley; or standing still at bus/train stops). Two coupled LSTM networks, Pedestrian movement LSTMs (one per target) and the corresponding Scene-LSTMs (one per grid-cell) are trained simultaneously to predict the next movements. We show that such common path information greatly influences prediction of future movement. We further design a scene data filter that holds important non-linear movement information. The scene data filter allows us to select the relevant parts of the information from the grid cells' memory relative to a target's state. We evaluate and compare two versions of our method with the Linear and several existing LSTM-based methods on five crowded video sequences from the UCY [1] and ETH [2] datasets. The results show that our method reduces the location displacement errors compared to related methods and specifically about 80% reduction compared to social interaction methods.*

## 1. Introduction

Human movement trajectory prediction is a challenging problem in computer vision. Given the past movement trajectories of pedestrians (targets) in a video sequence, the goal is to predict their future trajectories (lists of continuous two-dimensional locations). Human trajectory prediction has many real-world applications such as autonomous driving cars [1]: need to be able to predict the future locations of pedestrians on the street to avoid accidents; the robotic navigation systems [2, 3]: to help robots navigate through crowds by recognizing surrounding pedestrians and making movement decisions to avoid collisions; intelligent human tracking systems [4, 5]: capability to recognize and track all pedestrians in a scene.

For the most part, predicting future human trajectories is difficult. There are many possible future trajectories, especially in open areas (school yards, beaches, town squares, etc.) where people can move and change directions freely at any time. Social interactions can impact decisions of the next movements as well. For example, a group of people walking together in the past may tend to continue walking together in the near future. Structures can define specific paths within a scene. For instance, people walking out of an alley (Figure 1) tend to turn right/left to continue walking on sidewalks instead of going straight on to street. Designing a model that understands the scene context in conjunction with human movement model to help predict human trajectories accurately is both desirable and difficult.

To deal with these challenges, several methods [4, 7-13] have been proposed. The existing LSTM-based methods can be categorized into 2 types: social-interaction methods, which model social interactions among humans; and social-scene methods, which model social interactions and scene context simultaneously. These methods leverage the power of LSTM networks (a specific type of recurrent neural network (RNN)), that can characterize individual target's movement behaviors or social-interaction behaviors by using its memory cell. The memory cell is the key component leading the LSTM to solve many time-series data related problems such as machine translation [14], hand writing generation/recognition [15], and image generation [16]. The scene context is used in the social-scene methods [10, 13, 22] to complement the social features in order to improve the human movement prediction accuracy. These methods hypothesize that people tend to have the same walking patterns in similar scene layouts. To extract the scene features, [10, 22] utilize convolutional neural network (CNN) [17] that have been successfully used in image classifications [18, 19]. CNNs can produce similar scene features for new scene images and use them as inputs to model human movement behaviors.

In this paper, we propose and develop a scene model, called Scene-LSTM, where he scene is divided into equal-sized grid-cells which are further divided into sub-grids to provide more accurate spatial locations within the cell. Each grid-cell, assigned a memory cell, is modeled by a Scene-LSTM. The system learns human walking behaviors in grid-cells of the scene. The grid-cell memories (Scene-LSTM) are trained simultaneously with the pedestrian LSTMs. With the common human movements encoded in the grids' memory cells, the future human locations are forecasted more accurately. Figure 1 illustrates an example of our prediction results. The trained grid-cell memories (pink cells) are learned during the training steps. In the testing step, the trained scene information is used to correctly predict that the target will take a right turn moving out of the alley (a non-linear trajectory) and continue walking on the sidewalk.

The challenge is that in a given grid-cell, there can be multiple human trajectories possible (dissimilar walking patterns: different directions, velocities, and degrees of non-linearity). We implement a scene data filter (SDF), to let each target, based on its state, selects the relevant parts of grid cells' memories to predict its next location. The main components of SDF are a "hard filter" and a "soft filter". The grid cells will collect data based on non-linear movement information during the training stage. The goal of the "hard filter" is to allow the "non-linear" grid-cell memory to influence the prediction of human trajectory. The non-linear grid cells are those encompassing the non-linear human movements in the scene (e.g., a change of direction within the grid-cell). The soft filter then uses a target's current location, plus its state information as activations to select the relevant scene data from the hard filter. The final filtered scene data is then combined with the target's movement to predict this target's next location.

In summary, the contributions of this paper are:
- A new LSTM-based scene model is learned simultaneously with the traditional LSTM-based human walking model. We show the significant impact of the scene model on predicting human trajectories in the tested video sequences.
- We implement the Scene Data Filter (SDF) module to control the influence of each grid-cell using "hard-filter" for non-linearity stimulus, and "soft-filter" to select the scene data that predicts the target's trajectory.
- We model the walking behavior of each target using location offsets instead of absolute locations used in prior research [7]-[10]. Location offsets has shown to generate good results in hand writing generation [15], however it has not been applied to human trajectory prediction.

We define the human trajectory prediction problem in Section 1.1. In Section 2, we review the related work. Our model is described in Section 3. We present our results in Section 4 followed by conclusions and future work in Section 5.

## 1.1 Problem Definition

The problem under consideration is prediction of human movement trajectories in static crowded scenes. Let's define $X_i^t = (x_i^t, y_i^t)$ as the spatial location of target i at time t, and N as the number pedestrians in the number of observed frames $T_{obs}$. The problem can be formulated as: Given the trajectories of all targets in observed frames: $\{(x_i^t, y_i^t)\}$, where $t = 1, …, T_{obs}$ and $i = 1,2, …, N$, predict the next locations for each target in the number of predicted frames $T_{pred}$.

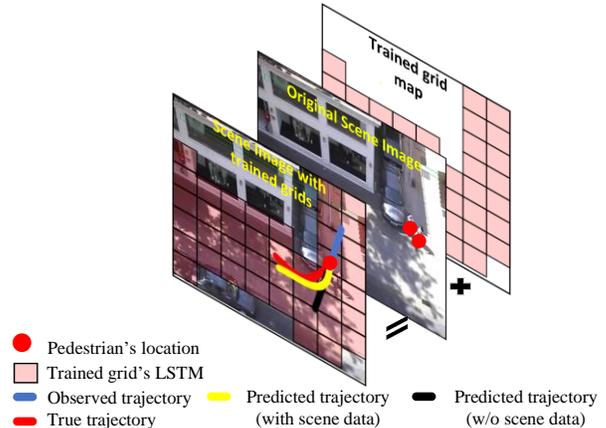

Figure 1: We learn the scene data by training our Scene-LSTM model for each grid-cell in a video sequence. The figure show that when using the trained grid cells (pink cells), the predicted trajectory (red) is more accurate than not using it (black).

## 2. Related Works

We classify the related research for human trajectory prediction into two categories: social interaction models and social-scene models.

**Social Interaction Models:** Some models of this type characterize the social interactions (e.g. grouping, avoiding, etc.) as features and calculate the next locations of each pedestrian by minimizing some function of these features. For example, [2] calculates the desired velocities of each pedestrian at the current frame by minimizing the energy function of collision avoidance, speed, and direction towards the pedestrians' final destinations. [20] extends the model in [2] with social group behaviors such as attractions and groupings using energy functions. The final energy function is minimized using gradient descent [27]. [4] characterizes human movements and collision potentials using Gaussian processes with multiple particles. Each particle represents a possible location in the next frame of a target. The maximum a posteriori (MAP) method is applied to minimize the collision potential to yield the best next locations of each target. Recently, several LSTM based methods have been proposed to model the social interactions. *Social-LSTM* [7] hypothesizes that the states (directions, speeds) of each pedestrian are influenced by the nearby pedestrians within a rectangle area. The current states of these neighboring pedestrians are used as inputs to calculate the target's next states. *Social-Attention* [8] uses the structural RNN network [21] to model the social interactions. The structural RNN network [21] is known for human activity detection and anticipation because it is capable of modeling spatiotemporal tasks. Different from [7], [8] considers the social interactions in the entire scene. Specifically, the faraway people may also have social impacts on the target's movements. These social impacts of other people on the main target are represented by an attention vector. This vector is calculated as a weighed sum of these people's current states and used as an input to predict the target's next locations.

**Social-Scene Models**: There is relatively a small body of recent work focused on studying the impacts of scene structures (e.g. buildings, static obstacles, etc.) on human trajectory prediction. These methods [10, 13, 22] combine scene features with social interactions [7] to predict human movement trajectories. [10] and [22] explore the scene influences by extracting features of the scene layouts using CNN and use them as inputs to the target's LSTM network to learn human walking behaviors. [13] hypothesizes that the closer each target is to the scene obstacles, the more impact they have on the target's walking behavior. They measure the distances between the targets and obstacles in the scene. The above methods do not characterize the relationship between the past trajectories and the current movements of a target at a specific area of a scene. In this work, we model these relationships by learning the past trajectories in each grid cell of a scene and let the scene's memory cell influence the target trajectories. We compare our method to the social interaction methods [7], [8] and the social-scene method [10].

## 3. System Design

We will present a brief review of LSTM networks followed by an overview of our system and its components.

### 3.1 LSTM: A Review

We present our introduction of LSTM in connection with the human trajectory prediction method introduced in [7]. LSTM network [23] is a class of recurrent neural networks (RNN). An LSTM consists of a memory cell $c$, an input gate $i$, an output gate $o$, and a forget gate $f$. The memory cell stores and remembers information (states) from the past time-series data. The input gate controls which new data flow into the memory cell. The output gate controls the "remind/remember" parts of data in the memory cell. The output gate controls which parts of data are used to calculate the output (i.e. the hidden state $h$). In human trajectory prediction, the memory cell $c$ is utilized to model the movement behavior of each target in a scene [7]. Given the hidden state $h_{t-1}^i$, memory cell $c_{t-1}^i$, and the current location $X_t^i = (x_i^t, y_i^t)$ of the target i, the LSTM-based network calculates the predicted locations $\hat{X}_t^i = (\hat{x}_t^i, \hat{y}_t^i)$ of each target i at time t as follows:

$$(h_t^i, c_t^i) = \text{LSTM}\big((h_{t-1}^i, c_{t-1}^i), (x_t^i, y_t^i); W\big) \quad (1)$$

$$(\hat{x}_t^i, \hat{y}_t^i) = W_{\hat{x}h} h_t^i + b_o \quad (2)$$

where W denotes the set of weight matrices of an LSTM unit. $W_{\hat{x}h}$ is the weight matrix between the hidden state $h_t^i$ and the output layer $\hat{X}_t^i$. $b_o$ is the bias vector of the output layer. The function LSTM(·) consists of following functions:

$$i_t = \sigma(W_{ix} x_t + W_{ih} h_{t-1} + W_{ic} c_{t-1} + b_i) \quad (3)$$
$$f_t = \sigma(W_{fx} x_t + W_{fh} h_{t-1} + W_{fc} c_{t-1} + b_f) \quad (4)$$
$$c_t = f_t c_{t-1} + i_t \tanh(W_{cx} x_t + W_{ch} h_{t-1}) + b_c \quad (5)$$
$$o_t = \sigma(W_{ox} x_t + W_{oh} h_{t-1} + W_{oc} c_t + b_o) \quad (6)$$
$$h_t = o_t \tanh(c_t) \quad (7)$$

where $i_t, f_t$ and $o_t$ denote the input gate, forget gate and output gate at time t. The matrix $W_{AB}$ denotes the weight matrix between the layer B and A (e.g. $W_{ix}$ is the weight matrix between the input layer $X_t$ and the input gate $i_t$). $\sigma(\cdot)$ denotes the sigmoid activation function and b terms denote the bias vector. We maintain that using LSTM network to model the targets' movement behavior independently is not sufficient because the target's movement behavior in the future may be different from the past and is highly dependent on the scene context. In the next sections, we will discuss the details of our Scene-LSTM model and demonstrate the influence the scene context has on human movement trajectories.

### 3.2 The overview of our model:

Figure 2 illustrates the overview of our system which consists of three main components: "pedestrian's movement", scene model, and the scene data filter (SDF). Before describing these components, we explain the pre-processing of the video images for input to the system.

**Pre-processing steps**: We first scale each scene image into 480x480 resolution and divide the resulting image into equal-sized grid cells (8x8 in our experiments) each to be modeled by a Scene-LSTM. We assign a memory cell $c_{gj}$ to cell $g_j$, where $g_j$ is the index of $j^{th}$ grid-cell of the scene (linearly indexed). Each grid cell is further divided into equal-sized sub-grids (4x4 in our experiments). Each target, i, is assigned a memory cell $c_i$, initialized with a zero vector for the pedestrian LSTM model. The grid-cells' memory values are updated during the training process. For each video sequence, we extract several batches of $T = T_{obs} + T_{pre}$ frames. We extract the trajectories of each target from each batch to be used as ground truth.

### 3.3 Pedestrian's movement

Given the current memory cell $c_i^t$, hidden state $h_i^t$ and spatial location $X_t^i = (x_i^t, y_i^t)$ of target i at time t, the "Pedestrian's movement" component calculates the predicted location $\hat{X}_t^i = (\hat{x}_t^i, \hat{y}_t^{t+1})$. Different from previous methods [7-10,13,22], we input the relative distance from the previous time step (location offset) ($\Delta x_i^t = x_i^t - x_i^{t-1}, \Delta y_i^t = y_i^t - y_i^{t-1}$) into this network component instead of the absolute values $(x_i^t, y_i^t)$. The absolute values $(x_i^t, y_i^t)$ are strongly biased to a specific scene layout which cause the network to fail to correctly predict the human trajectories in new scenes with a completely different scene layout. For example, if the network is trained on a scene that people mostly move horizontally, it will fail to predict vertical-movement trajectories in another scene during testing. On the other hand, the relative distance $(\Delta x_i^t, \Delta y_i^t)$, representing how far a target will move from the current location, is not dependent on the scene layout. Thus, by using the relative distance as input, the network is able to model the target's walking behavior more accurately. Given input $I_i^t = (\Delta x_i^t, \Delta y_i^t)$, we calculate the predicted relative location $(\Delta \hat{x}_i^t, \Delta \hat{y}_i^t)$ as follows:

$$e_i^t = \text{ReLU}(W_{ie} I_i^t) \tag{8}$$
$$(h_i^t, c_i^t) = \text{ReLU}(\text{LSTM}((h_i^{t-1}, c_i^{t-1}), e_i^t; W)) \tag{9}$$
$$[\mu_i^t, \sigma_i^t, p_t^i] = W_{of}[h_i^t + F_i^t] \tag{10}$$
$$(\Delta \hat{x}_i^t, \Delta \hat{y}_i^t) \sim N(\mu_i^t, \sigma_i^t, p_i^t) \tag{11}$$
$$(\hat{x}_i^t, \hat{y}_i^t) = (\hat{x}_i^{t-1} + \Delta \hat{x}_i^t, \hat{y}_i^{t-1} + \Delta \hat{y}_i^t) \tag{12}$$

where $e_i^t$ is the embedded vector (a dense vector representation of a target's location offset in a high dimensional space). $W_{ie}$ and $W_{of}$ are the weight matrices. $F_i^t$ is the filtered scene data of target $i$. $F_i^t$ has the same size as $h_i^{t+1}$ and will be described in Sections 3.4 and 3.5. The predicted relative location $(\Delta x_i^t, \Delta y_i^t)$ is estimated by the bivariate Gaussian distribution $N(\mu_i^t, \sigma_i^t, p_i^t)$, which has three parameters: the mean $\mu_i^t = (\mu_x^t, \mu_y^t)_i$, the standard deviation $\sigma_i^t = (\sigma_x^t, \sigma_y^t)_i$, and the correlation coefficient $p_i^t$. Finally, the predicted location of target $i$ at time $t$ is calculated in Equation (12).

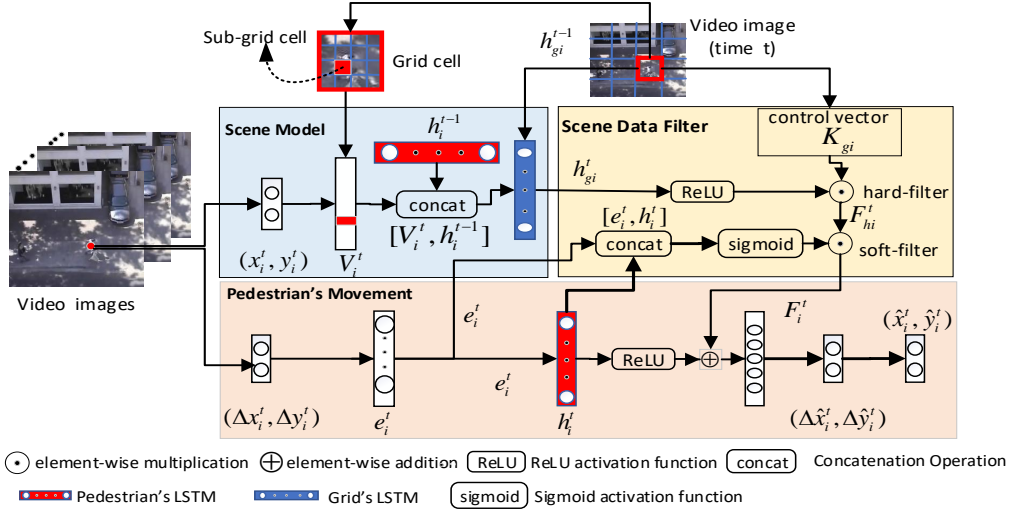

Figure 2: The overview of our model.

### 3.4 Scene Model

The "Scene Model" is responsible for training of the grid-cells' memories in order to characterize the common trajectories in the cells. The output of this component is the grid-cells' hidden states $h_{gi}^t$, which will then be filtered in the "Data Filtering" module to obtain $F_i^t$. The steps of calculating $h_{gi}^t$ are as follows:

- Given the spatial location of target $i$ $(x_i^t, y_i^t)$ at time $t$ (e.g. the red circle in the video images in Figure 2), a one-hot vector of location $V_i^t$ is calculated. The one-hot vector $V_i^t$ represents the relative location of this target to the grid cell, where this target walks in. $V_i^t$ is a vector of size 16 with values $[0, \ldots 1, \ldots 0]$, where 1 indicates which sub-grid within the current grid-cell the target occupied (the filled red square in the grid cell, Figure 2).
- Next, the concatenation of previous states of target $i$ $h_i^{t-1}$ and current one-hot vector of location $V_i^t$ is used as an input to calculate the state of this grid-cell:

$$(h_{gj}^t, c_{gj}^t) = \text{LSTM}\big((h_{gj}^{t-1}, c_{gj}^{t-1}), [V_i^t, h_i^{t-1}]; W_{gi}\big) \tag{13}$$

where $[V_i^t, h_i^{t-1}]$ denotes the concatenation operation of $V_i^t$ and $h_i^{t-1}$; $W_{gj}$ denotes the set of weight matrices in the LSTM network (the blue rectangle in Figure 2) of grid-cell $g_j$; $h_{gj}^{t-1}$ is the previous hidden state of grid-cell $g_j$. Initially, the hidden state $h_{gj}$ and the memory cell $c_{gj}$ of each grid-cell are set to zero vectors and their values are updated through training process. The hidden state $h_{gj}^t$, which carries the scene information for each target $i$, is used as an input to the Scene Data Filter (SDF) module. The SDF is responsible for selecting which parts of $h_{gj}^t$ will impact the target's movements.

Figure 3 shows an example of the trained grid cells (in pink color). These cells record the common human paths with respect to the scene structures or obstacles such as building (UCY-Zara01 in Figure 3a), unwalkable pavement with snow (ETH-Univ in Figure 3b), or trees (UCY-Univ in Figure 3c). Based on the common historic movements within a grid cell, the grid's state will help predict the human trajectories more accurately.

**Algorithm 1: Scene Data Filter**
**Input**: A grid's hidden state $h_{gj}^t$
A target's hidden state $h_i^t$
Embedded vector $e_i^t$
A set of non-linear trajectories $\mathcal{L}$
**Output**: Filtered data scene for target i  $F_i^t$
**Hard-filter:**
1. Define a hard control vector $K_{gj}$ for each grid.
   $K_{gj}$ is the same size as $h_i^{t+1}$.
2. If any human movement trajectory exists in $g_j \in \mathcal{L}$, then:
3.    $K_{gj} = [1, 1..1]$
4. else:
5.    $K_{gj} = [0, 0 ...,0]$
   // Output from hard-filter
6. $F_{hi}^t = relu(h_i^t) \odot \mathcal{K}_{gj}$
**Soft-filter:**
7. $F_i^t = \sigma([e_i^t, h_i^t]) \odot F_{hi}^t$
**Notation**: $\odot$ : element-wise multiplication.
  $\sigma(.)$: sigmoid activation function

## 3.5 Scene Data Filter (SDF)

When various trajectories co-inside within a grid cell as is often the case in open space or when the target can move freely without any scene constraints, the information learned by the grid-cell's memory can be chaotic and thus not helpful in predicting and adjusting a target's trajectory. To account for this, we design the "Scene Data Filter" (SDF) module that allows a target to select the relevant information depending on its state, $F_i^t$, from the hidden state $h_{gj}^t$ of where this target is walking, i.e., grid-cell $g_j$. Algorithm 1 describes the SDF computations. This module consists of two filters: a hard-filter and a soft-filter. The idea of using hard-filter is that we only allow the scene data flow from the non-linear grid-cells to influence the target's movement prediction. The non-linear grid-cells are those encompassing non-linear human trajectories learned during the training stage. The non-linear degree $\Phi_i$ of each target trajectory (length T) is calculated as follows:

$$\Phi_i = abs\left(\frac{y_T - y_0}{2} + y_0 - y_m\right) \quad (14)$$

where $y_o, y_m, y_T$ are y-axis locations at the beginning, the middle, and the end of a trajectory. The trajectory of target i is a non-linear trajectory if its $\Phi_i$ is greater than a defined threshold (0.2 in our experiments). We control the hard-filter of a grid cell using the control vector $K_{gj}$. The values of $K_{gj}$ are calculated as steps 1 to 5 in Algorithm 1. $K_{gj}$ is then used to calculate the filtered scene data $F_{hi}^t$ in step 6. The filtered scene data $F_{hi}^t$ from the hard filter is passed to the soft filter to calculate the final filtered scene data $F_i^t$. The soft filter is responsible for selecting and using a target's relevant portion of $F_{hi}^t$ as the final filtered scene data $F_i^t$. This filter computes $F_i^t$ by concatenating the embedded vector $e_i^t$ and hidden state $h_i^t$ as an activation (using sigmoid activation function).

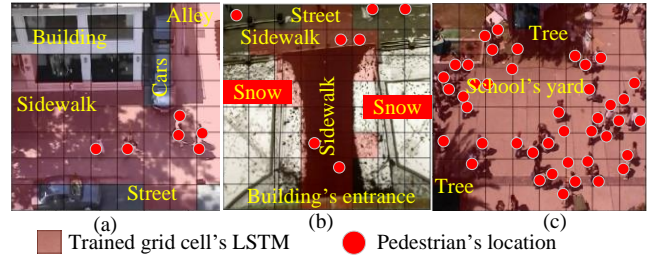

(a)   (b)   (c)
■ Trained grid cell's LSTM   ● Pedestrian's location

Figure 3: Example of trained grid cells (pink) of different video squences : (a) UCY-Zara01, (b) ETH-Univ, (c) UCY-Univ.

**Training Loss.** we train our model by minimizing the negative log-likelihood loss L [21]:

$$L(W) = -\sum_{i=0}^{N}\sum_{t=0}^{T} \log(P(x_i^t, y_i^t | \sigma_i^t, \mu_i^t, p_i^t)) \quad (15)$$

Where, W is the set of weight matrices. N is number of targets, $T = T_{obs} + T_{pred}$ is the number of frames used for training. $(x_i^t, y_i^t)$ is the true location of target i at time t. $\sigma_i^t, \mu_i^t, p_i^t$ are bivariate Gaussian parameters. By minimizing L(W), we maximize the likelihood that the predicted location $(\hat{x}_i^t, \hat{y}_i^t)$ is closer to the true location $O_i^{t+1} = (x_i^{t+1}, y_i^{t+1})$.

**Forecasting Trajectories (Testing).** During testing, we fit the trained model to the observed trajectories ($t = 0, .., T_{obs}$), calculate the targets' hidden states, and update the grid-cells' memories to be used next. We then use each target's location and hidden state at time $t = T_{obs}$ to forecast the next locations at time $t = T_{obs+1}, ..., T$.

## 4. Evaluation:

### 4.1 Datasets and Metrics

**Datasets**: As with the related prior research [7], [8], [10], we evaluate our model on two publicly available datasets: walking pedestrians dataset provided by ETH Zurich (ETH) [1] and crowd data provided by University of Cyprus (UCY) [2]. These datasets contain 5 video sequences (ETH-Hotel, ETH-Univ, UCY-Univ, ZARA-01, and ZARA-02) consisting of a total of 1536 pedestrians with different movement patterns. These sequences are recorded in 25 frames per second (fps) and contain 4 different scene backgrounds. UCY dataset only provides the annotated data (pedestrian's locations and identities) at control points, where people change walking directions; we linearly interpolate these locations at control points to get locations of each target in each frame.

**Metrics:** we evaluate our system using three metrics:
(a) *Average displacement error (ADE)*: The mean square error (MSE) over all locations of predicted trajectories and the true trajectories. The metric was first introduced in [2] and subsequently used in several reports [7], [8], [10].

$$ADE = \frac{\sum_{i=1}^{N}\sum_{t=0}^{t=T}\|\hat{X}_i^t - O_i^t\|_2}{N * T_{pred}} \quad (16)$$

where, $\hat{X}_i^t$ and $O_i^t$ are the predicted and the true locations of target $i$ at time t respectively. N is the number of targets and $T_{pred}$ is the predicted trajectory length.

(b) *Average non-linear displacement Error (NDE):* The average MSE over all locations of non-linear predicted trajectories and true trajectories; Equation (14).

(c) *Average Final Displacement Error (FDE):* The mean square error at the final predicted location and the final true location of all human trajectories.

$$\text{FDE} = \frac{\sum_{i=1}^{N} \|\hat{X}_i^T - O_i^T\|_2}{N} \quad (17)$$

Similar to [10], we report all prediction errors in the normalized range [-1,1] as the homograph matrix, used to covert location in pixel values to meters, is not publicly available for UCY dataset.

**Comparison with existing methods**: We compare results of our models with the following methods:

- Linear model [8], [9]: uses a linear regressor to estimate the linear parameters by minimizing the mean square error, assumes that pedestrians move linearly.
- LSTM [7]: models pedestrian's states without considering social interactions or scene information.
- Social-LSTM [7]: models the social interactions between pedestrians using "social" pooling layers. As the authors' code is not made available, we implemented it by referencing the code given in [8]. Different from [8], we extract batches of frames consisting of one or many human trajectories per time step. However, [8] skips several frames by jumping to other time steps randomly. Our batch extraction method includes more data, reduces overfitting problem, and hence produces better results than [8].
- Social-Attention [8]: uses the structural neural network and attention module to model social interactions. We use the author's publicly available code to generate results for Social-Attention.
- SS-LSTM [10]: uses both scene information and social interactions to model human movement behaviors. We use the results reported in this paper to compare with our method as their code is not available

We report on two variants of our method: (a) Scene-LSTM-a uses scene data from all grid cells to predict human movements, and (b) Scene-LSTM-n only uses scene data from the non-linear grid cells. We do not compare with result with [9, 12] as there is no publicly available implementation, and because the reported results are stated in metrics different than other reported research. We intend to make our implementation available upon publication.

### 4.2 Implementation details:

The implementation is doe in PyTorch deep learning framework [25]. The size of all memory cells and hidden state vectors is set to 128. We use an 8x8 grid for video scene and 4x4 sub-grid for grid-cells. The network is trained with Adam optimizer [26], an extension to stochastic gradient descent, to update network weights during the training process. The learning rate is 0.003, and the dropout value is 0.2. The value of the global norm of gradients is clipped at 10 to ensure stable training. The model is trained on GPU Tesla P100-SXM2

**Training**. The training stage is carried out in two stages:

*Stage 1*: Indexing the five video sequences (ETH-Univ, UCY-Univ, UCY-Zara01, and UCY-Zara02) as (i, j, k, l, m), we train and validate four video sequences (Vi, Vj, Vk, Vl) and select the best set of weights generated from this stage (lowest ADE) to be used in stage 2 for the remaining (unseen) video sequence Vm; this process is referred to "leave-one-out". This process is repeated five times for five permutations of (i, j, k, l, m) to train/validate four videos in stage 1 in order to obtain five set of weights to be used in stage 2 for each of the remaining unseen videos. For each permutation, we split the data of the four training videos into 80% used for training and 20% for validation steps. The training is carried out for 100 epochs (the number of times the entire training data is used once, before weights are updated).

*Stage 2*: The best trained set of weights for each of the five permutations obtained from stage 1, is used to train the fifth unseen video. During this stage, 50% of the fifth video is used for training which is carried out for 10 epochs. During this training step, we learn the new scene information. The best set of weights from this process is then used for testing the remaining 50% of this video sequence.

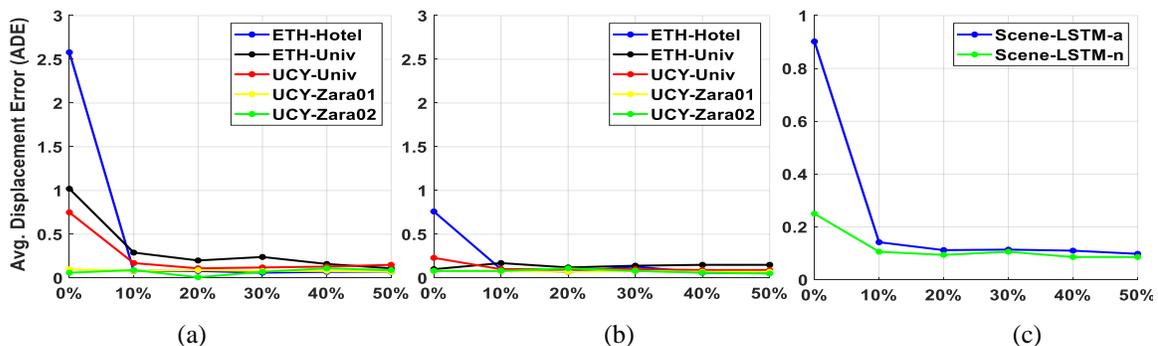

Figure 4: The impact of the amount of training data in the stage-2 on the prediction error (ADE) for each video sequences using all-grid model in: *Scene-LSTM-a* (a), nonlinear-grid model: *Scene-LSTM-n* (b), and comparision between *Scene-LSTM-a* and *Scene-LSTM-n* (c); note there is scale difference in plot (c).

**Testing**: The system is initialized with the final set of weights from stage 2. The testing process is repeated at each frame by observing eight frames (3.2 seconds) and predicting the next 12 frames (4.8 seconds) in a sliding window fashion.

We have implemented the same training and testing procedures for all the methods we compare our results to (Linear [9], LSTM [7], Social-LSTM [7], Attention-LSTM [8]) so the comparisons are as similar as possible. The exception is SS-LSTM [10] (implementation not available) where, we use the results reported in their paper. It uses a training process similar to Stage 1.

### 4.3 The impacts of training data amount in *stage-2*

In this section, we explore the impact of the amount of data used for training the new (unseen) video in *stage-2* on the testing accuracy of our two models: *Scene-LSTM-a* and *Scene-LSTM-n*. Figure 4 shows the testing error (ADE) as a function of ranging the amount of training data from 0% to 50% in *stage-2*. The testing is always carried out on the last 50% of the video. In Figure 4.c, we observe that the average prediction errors are significantly reduced for both models as the amount of training data increases. The ADE error is reduced by 89% and 68% in *Scene-LSTM-a* and *Scene-LSTM-n* respectively at 50% training data compared to 0%. Figure 4, a and b show that there are some video sequences (e.g. UCY-Zara01 and UCY-Zara02) that only need a small amount of scene information through observation trajectories to achieve good results, while the other video sequences (e.g. ETH-Hotel, ETH-Univ, UCY-Univ) require more of the scene information to perform better. This is because the scene structures of ETH-Hotel, ETH-Univ, UCY-Univ are different from each other and from the scene in UCY-Zara01 and UCY-Zara02 that have the same layout with the advantage of the similar common human movements that have been learned during *stage-1*.

### 4.4 Comparison with related methods

**Quantitative Results**. We compare the results of our models with five existing methods. The testing results of each sequence are calculated on the last 50% data of each video sequence. The quantitative results in Table 1 show that the two versions of our model: *Scene-LSTM-a* and *Scene-LSTM-n*, outperform the others on all three metrics on most video sequences. In Scene-LSTM-a the prediction error is significantly reduced compared to social interaction methods (i.e. by 68% from Social-LSTM and 83% from Attention-LSTM). We achieve better ADE, NDE in most video sequences and better FDE in all sequences compared to social-scene method SS-LSTM [10]. Results in Table 1 confirm that our scene model is more efficient in predicting the final locations (FDE) of each target than SS-LSTM. Note that SS-LSTM uses both scene and social features and results reported in Table 1 are those available from their published paper (implementation is not available). We came to the same conclusion reported in [9] that the Attention-LSTM and Social LSTM do not perform as well as the LSTM model.

| Metrics | Sequences | Linear | LSTM[7] | Social LSTM[7] | Social Attention[8] | SS-LSTM [10] | Scene-LSTM-a (Our method) | Scene-LSTM-n (Our method) |
|---|---|---|---|---|---|---|---|---|
| Average displacement error (ADE) | ETH-Hotel | 0.21 | 0.21 | 0.25 | 0.44 | 0.07 | **0.06** | **0.06** |
| | ETH-Univ | 0.17 | 0.16 | 0.18 | 0.44 | **0.10** | 0.11 | **0.10** |
| | UCY-Univ | 0.23 | 0.27 | 0.25 | 0.19 | **0.08** | 0.11 | 0.09 |
| | UCY-Zara01 | 0.36 | 0.33 | 0.37 | 0.47 | **0.05** | 0.07 | 0.07 |
| | UCY-Zara02 | 0.15 | 0.19 | 0.19 | 0.56 | **0.05** | 0.06 | **0.05** |
| | **Average** | 0.22 | 0.23 | 0.25 | 0.42 | **0.07** | 0.08 | **0.07** |
| Average non-linear displacement error (NDE) | ETH-Hotel | 0.18 | 0.21 | 0.19 | - | - | **0.07** | 0.08 |
| | ETH-Univ | 0.22 | 0.19 | 0.21 | - | - | 0.16 | **0.13** |
| | UCY-Univ | 0.23 | 0.28 | 0.25 | - | - | 0.11 | **0.10** |
| | UCY-Zara01 | 0.32 | 0.33 | 0.29 | - | - | **0.09** | **0.09** |
| | UCY-Zara02 | 0.17 | 0.21 | 0.21 | - | - | 0.07 | **0.06** |
| | **Average** | 0.22 | 0.24 | 0.23 | - | - | 0.10 | **0.09** |
| Final displacement error (FDE) | ETH-Hotel | 0.25 | 0.23 | 0.29 | 0.55 | 0.12 | **0.06** | 0.07 |
| | ETH-Univ | 0.24 | 0.29 | 0.34 | 0.50 | 0.24 | 0.19 | **0.18** |
| | UCY-Univ | 0.03 | 0.05 | 0.03 | 0.10 | 0.13 | 0.02 | **0.02** |
| | UCY-Zara01 | 0.32 | 0.28 | 0.32 | 0.62 | 0.08 | 0.08 | **0.07** |
| | UCY-Zara02 | 0.09 | 0.21 | 0.10 | 0.71 | 0.08 | 0.03 | **0.02** |
| | **Average** | 0.19 | 0.21 | 0.22 | 0.50 | 0.09 | 0.08 | **0.07** |
| Features | | I | I | So+I | So+I | So+Sc+I | Sc+I | Sc+I |

Table 1. The quantitative results on 5 different videos sequences. All methods predict human trajectories in 12 frames and using 8 observed frames. I denotes Individual movement. So denotes Social Interactions. Sc denotes Scene Information.

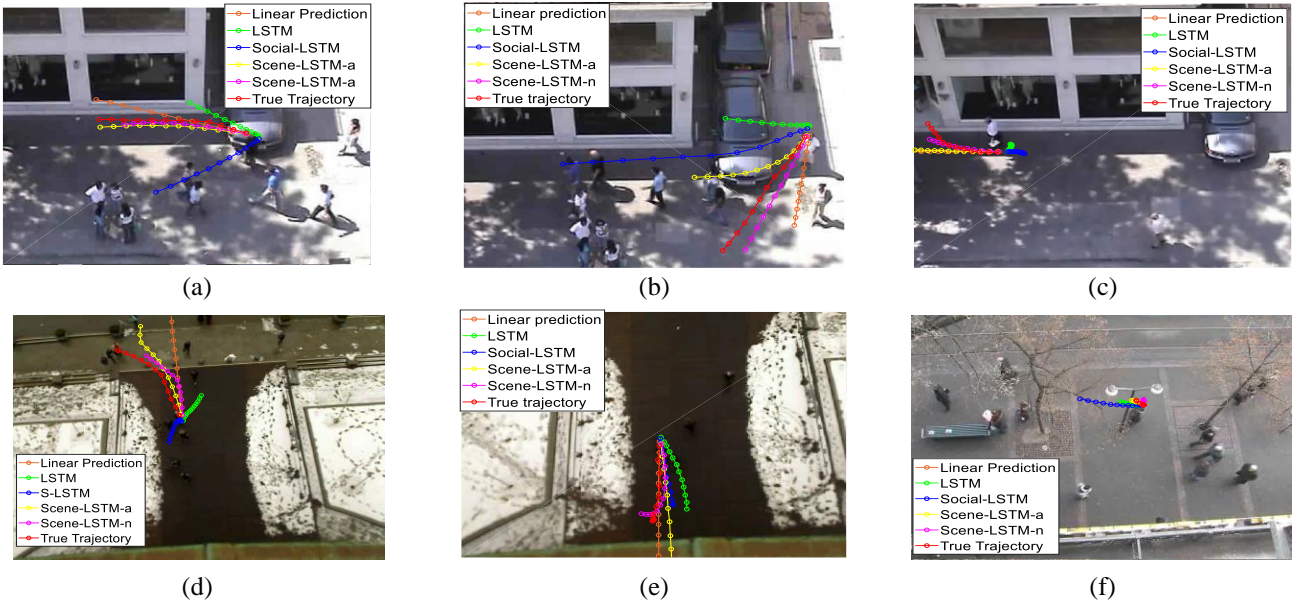

Figure 5: Some examples of predicted trajectories from different methods in various video sequences (a,b) UCY-Zara01, (c) UCY-Zara01, (d,e) ETH-Univ, (f), ETH-Hotel. The scene images are zoomed out for better visualization.

*Comparison between the two variants of our model: Scene-LSTM-a and Scene-LSTM-n:* Scene-LSTM-n generates better results for all video sequences where non-linear movements are observed. The one exception is in ETH-Hotel sequence where most people move linearly and very few non-linear grid cells influence the prediction. Our results and observations lead us to conclude that *Scene-LSTM-a* does not predict the actual scenarios best in scenes with non-linear movements. Using selected grid cells, as in non-linear grids in *Scene-LSTM-n* to influence the prediction, generates more accurate trajectories.

**Qualitative Results.** Figure 5 shows examples of the predicted human trajectories from five methods plus ground truth. Figure 5a shows that both our models correctly predict the trajectories of the target to move to the building shadow. The linear model and LSTM models predict this target to go into either the car or inside the building. We predict more accurate trajectories because throughout the training process of Scene-LSTMs, the grid cells in the position in which the target is standing have seen many simmilar paths where people often try to avoid the sunlight and go under the building's shadow. In Figure 5b, the *Scene-LSTM-n* model produces the best result and closest to the true trajectory. In this scenario, the target is leaving the alley to turn left and walk on the sidewalk (correct prediction). The linear model predicts this target to go straight out to the street. *Scene-LSTM-a* model incorrectly produces a sharp non-linear trajectory because all the grid cells in that area are trained with horizontal movements according to the common human movments on the sidewalk. A similliar senario is observed in Figure 5d, when most people going out of the alley, turn left on the sidewalk. In Figure 5c, we obtain accurate results where people often turn right from the sidewalk to enter the building, while all the other methods make incorrect predictions. Figure 5e and f are from videos (ETH-Univ and ETH-Hotel) where people tend to stop (standing around, Figure 5-f) or slow down (to open the door, Figure 5-e) at different areas of the scene. *Scene-LSTM-n* incorporates these movement behavior patterns more accurately than others.

## 5. Conclusion

In this work, we present two novel Scene-LSTM models for predicting human trajectories. We show that the scene information has significant impact in predicting on how people move. We have characterized the scene information by assigning and learning memory cells for each grid cell in a scene. The memory cell is capable to remember the useful information about how people moved in the past in a grid cell and use that information to predict the future trajectories. Our results show our methods outperform the existing methods.

In future work, we will investigate fusing the scene model with social model to improve prediction quality. We intend to further explore the social interactions not only among humans but also between human and other static or moving objects. Our goal is to apply our human trajectory prediction method to solve computer vision problems such as multi-target and multi-camera multi-target tracking systems.

# 6. References

[1] A. Lerner, Y. Chrysanthou, and D. Lischinski. Crowds by Example. Computer Graphics Forum, pages 655–664, 2007.

[2] S. Pellegrini, A. Ess, K. Schindler, and L. van Gool. You'll never walk alone: Modeling social behavior for multi-target tracking. In IEEE 12th International Conference on Computer Vision (ICCV), pages 261–268, 2009.

[3] R. P. D. Vivacqua, M. Bertozzi, P. Cerri, F. N. Martins, and R. F. Vassallo. Self-Localization Based on Visual Lane Marking Maps: An Accurate Low-Cost Approach for Autonomous Driving. IEEE Transactions on Intelligent Transportation Systems, pages 582–597, 2018.

[4] P. Trautman and A. Krause. Unfreezing the robot: Navigation in dense, interacting crowds. In International Conference on Intelligent Robots and Systems, pages 797–803, 2010.

[5] J. J. Leonard and H. F. Durrant-Whyte. Application of multi-target tracking to sonar-based mobile robot navigation. In 29th IEEE Conference on Decision and Control, pages 3118–3123, 1990.

[6] I. Ali and M. N. Dailey. Multiple human tracking in high-density crowds. Image and Vision Computing, pages 966–977, 2012.

[7] A. Alahi, K. Goel, V. Ramanathan, A. Robicquet, L. Fei-Fei, and S. Savarese. Social LSTM: Human Trajectory Prediction in Crowded Spaces. In IEEE Conference on Computer Vision and Pattern Recognition (CVPR), pages 961–971, 2016.

[8] A. Vemula, K. Muelling, and J. Oh. Social Attention: Modeling Attention in Human Crowds. arXiv:1710.04689 [cs], 2017.

[9] A. Gupta, J. Johnson, L. Fei-Fei, S. Savarese, and A. Alahi. Social GAN: Socially Acceptable Trajectories with Generative Adversarial Networks. arXiv:1803.10892 [cs], 2018.

[10] H. Xue, D. Q. Huynh, and M. Reynolds. SS-LSTM: A Hierarchical LSTM Model for Pedestrian Trajectory Prediction, In IEEE Winter Conference on Applications of Computer Vision (WACV), pages 1186–1194, 2018.

[11] T. Fernando, S. Denman, S. Sridharan, and C. Fookes, "Soft + Hardwired Attention: An LSTM Framework for Human Trajectory Prediction and Abnormal Event Detection," arXiv:1702.05552 [cs], Feb. 2017.

[12] A. Robicquet, A. Sadeghian, A. Alahi, and S. Savarese. Learning Social Etiquette: Human Trajectory Understanding In Crowded Scenes. In ECCV, pages 549–565, 2016.

[13] F. Bartoli, G. Lisanti, L. Ballan, and A. Del Bimbo, "Context-Aware Trajectory Prediction," arXiv:1705.02503 [cs], May 2017.

[14] I. Sutskever, O. Vinyals, and Q. V. Le, "Sequence to Sequence Learning with Neural Networks," arXiv:1409.3215 [cs], Sep. 2014.

[15] A. Graves, "Generating Sequences With Recurrent Neural Networks," arXiv:1308.0850 [cs], Aug. 2013.

[16] K. Gregor, I. Danihelka, A. Graves, D. J. Rezende, and D. Wierstra, "DRAW: A Recurrent Neural Network For Image Generation," arXiv:1502.04623 [cs], Feb. 2015.

[17] A. Krizhevsky, I. Sutskever, and G. E. Hinton. ImageNet Classification with Deep Convolutional Neural Networks. In Advances in Neural Information Processing Systems, pages 1097–1105, 2012.

[18] D. C. Cireşan, U. Meier, J. Masci, L. M. Gambardella, and J. Schmidhuber. Flexible, High Performance Convolutional Neural Networks for Image Classification. In the 22nd International Joint Conference on Artificial Intelligence (IJCAI), pages 1237–1242, 2011.

[19] Jurgen Schmidhuber. Multi-column deep neural networks for image classification. In CVPR, pages 3642-3649, 2012.

[20] K. Yamaguchi, A. C. Berg, L. E. Ortiz, and T. L. Berg. Who are you with and where are you going?. In CVPR, pages 1345–1352, 2011.

[27] L. Bottou. Large-Scale Machine Learning with Stochastic Gradient Descent. In Proceedings of COMPSTAT, pages 177–186, 2010.

[21] A. Jain, A. R. Zamir, S. Savarese, and A. Saxena. Structural-RNN: Deep Learning on Spatio-Temporal Graphs, pages 5308–5317, CVPR 2016.

[22] D. Varshneya and G. Srinivasaraghavan. Human Trajectory Prediction using Spatially aware Deep Attention Models. arXiv:1705.09436 [cs], May 2017.

[23] S. Hochreiter and J. Schmidhuber. Long Short-Term Memory. Neural Computation, pages 1735–1780. 1997.

[24] A. Graves. Supervised Sequence Labelling. In Supervised Sequence Labelling with Recurrent Neural Networks, pages 5–13, 2012

[25] "PyTorch." [Online]. Available: https://pytorch.org/. [Accessed: 23-Jun-2018].

[26] D. P. Kingma and J. Ba. Adam: A Method for Stochastic Optimization. arXiv preprint arXiv:1412.6980, 2014.